\def\year{2022}\relax
\documentclass[letterpaper]{article} 
\usepackage{aaai22}  
\usepackage{times}  
\usepackage{helvet}  
\usepackage{courier}  
\usepackage[hyphens]{url}  
\usepackage{graphicx} 
\usepackage{natbib}  
\usepackage{caption} 
\DeclareCaptionStyle{ruled}{labelfont=normalfont,labelsep=colon,strut=off} 
\usepackage{algorithm}
\usepackage{algorithmic}

\usepackage{makecell}
\usepackage{multirow}

\usepackage{amsmath}
\usepackage{array,tabularx}
\usepackage{lipsum}

\newenvironment{conditions*}
  {\par\vspace{\abovedisplayskip}\noindent
   \tabularx{\columnwidth}{>{$}l<{$} @{}>{${}}c<{{}$}@{} >{\raggedright\arraybackslash}X}}
  {\endtabularx\par\vspace{\belowdisplayskip}}

\usepackage{newfloat}
\usepackage{listings}
\floatstyle{ruled}
\newfloat{listing}{tb}{lst}{}
\floatname{listing}{Listing}
\title{Same Words, Different Meanings: Semantic Polarization in Broadcast Media Language Forecasts Polarization on Social Media Discourse}
\author {
    Xiaohan Ding,\textsuperscript{\rm 1}
    Mike Horning, \textsuperscript{\rm 2}
    Eugenia H. Rho \textsuperscript{\rm 1} \\
}
\affiliations {
    \textsuperscript{\rm 1} Department of Computer Science, Virginia Tech \\
    \textsuperscript{\rm 2} Department of Communication, Virginia Tech \\
    xiaohan@vt.edu, mhorning@vt.edu, eugenia@vt.edu
}

\usepackage{bibentry}

\begin{document}

\maketitle

\begin{abstract}
With the growth of online news over the past decade, empirical studies on political discourse and news consumption have focused on the phenomenon of filter bubbles and echo chambers. Yet recently, scholars have revealed limited evidence around the impact of such phenomenon, leading some to argue that partisan segregation across news audiences cannot be fully explained by online news consumption alone and that the role of traditional legacy media may be as salient in polarizing public discourse around current events. In this work, we expand the scope of analysis to include both online and more traditional media by investigating the relationship between broadcast news media language and social media discourse. By analyzing a decade’s worth of closed captions (2.1 million speaker turns) from CNN and Fox News along with topically corresponding discourse from Twitter, we provide a novel framework for measuring semantic polarization between America’s two major broadcast networks to demonstrate how semantic polarization between these outlets has evolved (Study 1), peaked (Study 2) and influenced partisan discussions on Twitter (Study 3) across the last decade. Our results demonstrate a sharp increase in polarization in how topically important keywords are discussed between the two channels, especially after 2016, with overall highest peaks occurring in 2020. The two stations discuss identical topics in drastically distinct contexts in 2020, to the extent that there is barely any linguistic overlap in how identical keywords are contextually discussed. Further, we demonstrate at-scale, how such partisan division in broadcast media language significantly shapes semantic polarity trends on Twitter (and vice-versa), empirically linking for the first time, how online discussions are influenced by televised media. We show how the language characterizing opposing media narratives about similar news events on TV can increase levels of partisan discourse online. To this end, our work has implications for how media polarization on TV plays a significant role in impeding rather than supporting online democratic discourse.\end{abstract}

\section{Introduction}
Mass media plays a vital role in democratic processes by influencing how institutions operate, political leaders communicate, and most importantly, how citizens engage in politics \cite{mcleod_community_1999}. Although it is no surprise that America’s two political divides speak different languages \cite{westfall_perceiving_2015}, research has also shown that partisan language in news media has sharply increased in recent years, particularly in broadcast news \cite{horning_pudit_2018}. This is concerning given that news consumption is critical for helping the public understand the events around them. According to Agenda Setting Theory, the language used by the media to frame and present current events impacts how the public perceives what issues are important \cite{mccombs_building_1997, russell_neuman_dynamics_2014}. 
\newline
\indent While some may have the impression that mainstream legacy media is decreasing in relevancy amid the explosive growth of online news via websites and social media, American news consumption is still overwhelmingly from television, accounting for nearly five times as much as online news consumption across the public \cite{allen_evaluating_2020}. Despite the notion that TV news consumption is more “passive” than reading the news, research shows that people tend to recall televised news better than online news \cite{eveland_learning_2002}. Further, a recent study comparing TV vs. internet news consumption found that there are four times as many Americans who are partisan-segregated via TV than via online news. In fact, TV news audiences are several times more likely to maintain their partisan news diets overtime, and are much narrower in their sources while even partisan online news readers tend to consume from a variety of sources \cite{muise_quantifying_2022}.
\newline
\indent
Yet studies on media polarization and ensuing public discourse are overwhelmingly based on online content \cite{garimella_political_2021}. For example, even research that analyzes data from traditional news outlets, solely relies on tweets from the official Twitter accounts of newspapers, TV shows, and radio programs rather than the direct transcription of content from these legacy media sources \cite{recuero_hyperpartisanship_2020}. This is due to the fact that unlike online information, legacy media data (e.g., closed captions) are harder to collect, exist in formats incompatible for quick pre-processing (e.g., srt files), and scattered across institutions that lack incentives to share data with academics. Hence, much of how mainstream legacy media affects online discourse is unknown.
\newline
\indent
In that sense, our analysis of a decade’s worth of closed captions from 24hr-broadcast TV news programs from America’s two largest news stations presents a unique opportunity to empirically demonstrate how linguistic polarization in broadcast media has evolved over time, and how it has impacted social media discourse. In this work, we examine how semantic differences in broadcast media language have evolved over the last $11$ years between CNN and Fox News (Study 1), what words are characteristic of the semantic polarity peaks in broadcast media language (Study 2), whether semantic polarity in TV news language forecasts polarization trends in social media discourse, and how language plays a role in driving relational patterns from one to the other (Study 3).
\newline
\indent In Study 1, we leverage techniques in natural language processing (NLP) to develop a method that quantitatively captures how semantic polarization between CNN and Fox News has evolved from 2010 to 2020 by calculating the semantic polarity of how socially important, yet politically divided topics (\textit{racism}, \textit{B\textit{lack} Lives Matter}, \textit{police}, \textit{immigration}, \textit{climate change}, and \textit{health care}) are discussed by the two news channels. We then use a model interpretation technique in deep learning to linguistically unpack what may be driving these spikes by extracting contextual tokens that are most predictive of how each station discusses topical keywords in 2020 (Study 2). To investigate whether partisan trends in broadcast media language influence polarity patterns in social media discourse, we use Granger causality to test whether and how semantic polarization between the two TV news stations forecasts polarization across Twitter audiences replying to @CNN and @FoxNews (Study 3). Finally to understand the language that drives the Granger-causal relations in how semantic polarity trends in televised news affect that across Twitter users (and vice-versa), we identify tokens that are most predictive of how topical keywords are discussed on TV vs. Twitter, separated by lag lengths that correspond to Granger-causality significance. Our contributions are as follows:
\begin{itemize}
 
    \item  We provide a novel framework for quantifying semantic polarization between two entities by considering the temporality associated with how semantic polarization evolves over time. Prior research that quantifies polarization as an aggregate measure from a single longitudinal data dump often leaves out key temporal dynamics and contexts around how polarity unfolds across time. Our framework incorporates temporal fluctuations by computing diachronic shifts using contextual word embeddings with temporal features.
    
    \item In showing how semantic polarization in broadcast media has evolved over the last 11 years, we go beyond providing a mere quantification of polarization as a metric by using Integrated Gradients to identify attributive tokens as a proxy to understand the contextual language that drives the 2020 ascent in semantic polarity between the two news stations. 
    
    \item We address the question of whether and how polarization in televised news language forecasts semantic polarity trends across Twitter, providing new evidence around how online audiences are shaped in their discourse by TV news language --- an important link that has not been empirically established at-scale in prior research.

    \item Finally, we use model interpretation to extract lexical features from different entities, to show which words drive significant Granger-causal patterns in how broadcast media language shapes Twitter discourse and vice-versa, thereby highlighting the manner in which language plays a key role in driving semantic polarity relations between online discussions and broadcast media language. 
\end{itemize}

\indent Our findings are one of the first to quantify how language characterizing opposing media narratives about similar news events on TV can increase levels of partisan discourse online. Results from this work lend support to recent scholarship in communications research, which theorizes that both media and public agendas can influence each other, and that such dynamics can polarize the manner in which the public engages in discourse, thereby influencing democratic decision-making at-large.
\section{Related Work}
\subsection{Temporal Dynamics in Linguistic Polarization}
Scholars have measured linguistic polarization through stance \cite{dash_divided_2022}, toxicity \cite{sap_risk_2019}, sentiment detection \cite{yang_quantifying_2017}, topic modeling \cite{ebeling_analysis_2022}, and lexicon-dictionaries \cite{polignano_hybrid_2022}. Such methods typically capture an aggregated snapshot of polarization across large textual corpora, providing a static quantification of media bias based on preset criteria. As a result, such approaches seldom capture temporal fluctuations in semantic polarity. Only a few research to date quantify linguistic polarization over time. For example, researchers have measured partisan trends in congressional speech by defining polarization as the expected posterior probability that a neutral observer would infer a speaker's political party from a randomly selected utterance \cite{gentzkow_measuring_2019} across various issues. Demszky et al. use this Bayesian approach to capture how social media discussions on mass shootings become increasingly divisive \shortcite{demszky_goemotions_2020}. 

Such methods, however, capture linguistic polarization as a general metric across nearly \textit{all} aggregated words within a textual corpus. Given the powerful salience of certain topical keywords that are inseparable from American politics, unlike prior work, we focus on how \textit{specific} keywords polarize over time. We depart from previous approaches by devising a framework that captures semantic polarity as a function of how two different entities semantically diverge across time in their contextual use of identical words that are central to the American public discourse. Specifically, we measure linguistic polarization by capturing the diachronic shifts in how two news stations use the same topical words on their news programs over an 11-year period. 

Further, in effort to understand the linguistic drivers behind how and why semantic polarization evolves, we aim to identify the source of diachronic shifts based on how two different entities use identical, politically contentious keywords. Few research has endeavored to detect causal factors underlying how the meaning of words changes over time, which remains an open challenge in NLP scholarship \cite{kutuzov_diachronic_2018}. Hamilton et al. show how word meanings change between consecutive decades due to cultural shifts (e.g., change in the meaning of “cell" driven by technological advancement: prison cell vs. cell phone) \shortcite{hamilton_cultural_2016}. Yet such research typically makes a priori assumptions that the language between two textual corpora of comparison are significantly related without statistically demonstrating how so. On the other hand, research that do provide evidence of statistical relations between two separate textual data \cite{dutta_measuring_2018}, generally do not explain what words drive the direction of semantic influence from one corpus to another. We overcome these limitations: first, we statistically validate the existence of a significant semantic relationship between two textual corpora, TV news language and social media discourse, through Granger-causality (Study 3). Then, we strategically separate our data by lag-times associated with significant Granger-causal relations and apply a deep learning interpretation technique to demonstrate which contextual words from TV news language influences Twitter discussions around specific topical keywords (and vice-versa) over time. 

\subsection{Who Influences Who? Setting a Media Agenda}
On any given day, newsrooms select news on a wide range of topics that cover a variety of issues. According to Agenda Setting Theory, this selection process leads audiences to see certain issues as more significant than others \cite{mccombs_building_1997}. Agenda Setting was particularly useful in decades past when media was much more consolidated. In the U.S. for example, three major networks (NBC, ABC, and CBS) provided the bulk of broadcast news content, and so their news agendas had broad influence on what audiences saw as important. However, today's media landscape is much more fragmented. People receive their news from any number of sources including television, newspapers, radio, online websites, and social media. In such an environment, Agenda Setting has certain limitations as the theory assumes that audiences are relatively passive and simply receive information from the media. This was true when news flowed one-way from the media to the mass public. 
\begin{table}[b]
\centering
\resizebox{1.0\columnwidth}{!}{
\begin{tabular}{|c||l|l|}
\hline  \multicolumn{2}{|c|}{Topic} & \multicolumn{1}{|c|}{Topical Keywords} \\
\hline 1&Racism &racism, racist \\
\hline 2&Black Lives Matter &blacklivesmatter \\
\hline 3&Police &police \\
\hline 4&Immigration &immigration, immigrants \\
\hline 5&Climate Change&climate change, global warming \\
\hline 6&Health Care &health care \\
\hline
\end{tabular} }
\caption{Our data consists of speaker turns that contain 9 target keywords pertaining to six core topics.}
\label{table1}
\end{table}

By contrast nowadays, the two-way form of communication afforded by the internet and social media makes it possible for the public to both influence media agendas and be influenced by them \cite{papacharissi2009journalism, barnard2018citizens}. As a result, new scholarship has argued through Intermedia Agenda Setting Theory that media agendas do not come entirely from within their own organizations, but from two other sources: other media and the mass audience. In the latter case, for example, audiences themselves can influence media agendas through online affordances (e.g., retweeting, commenting, sharing), to raise certain issues to prominence online \cite{rogstad2016twitter}. In this work, we explore these new emerging dynamics with an interest in not only understanding how news agendas are constructed, but how the possible adoption of different agendas might influence agenda dynamics as media interact with the public online.

\section{Study 1: Evolution of Semantic Polarity in Broadcast Media Language (2010-2020)}
NLP researchers studying diachronic shifts have been leveraging word embeddings from language models to understand how the meaning of a given word changes over time \cite{kutuzov_diachronic_2018}. In Study 1, we adapt methodological intuitions from such prior work to calculate \textbf{how semantic polarization --- \textit{the semantic distance between how two entities contextually use an identical word} --- evolves over time}. While scholars have examined online polarization in the context of understanding partisan framing and public discourse around current events, our work is the first to quantify and capture the evolution of semantic polarization in broadcast media language. 
\subsection{Dataset}
\begin{figure}[ht]
\centering
\includegraphics[width=0.97\columnwidth]{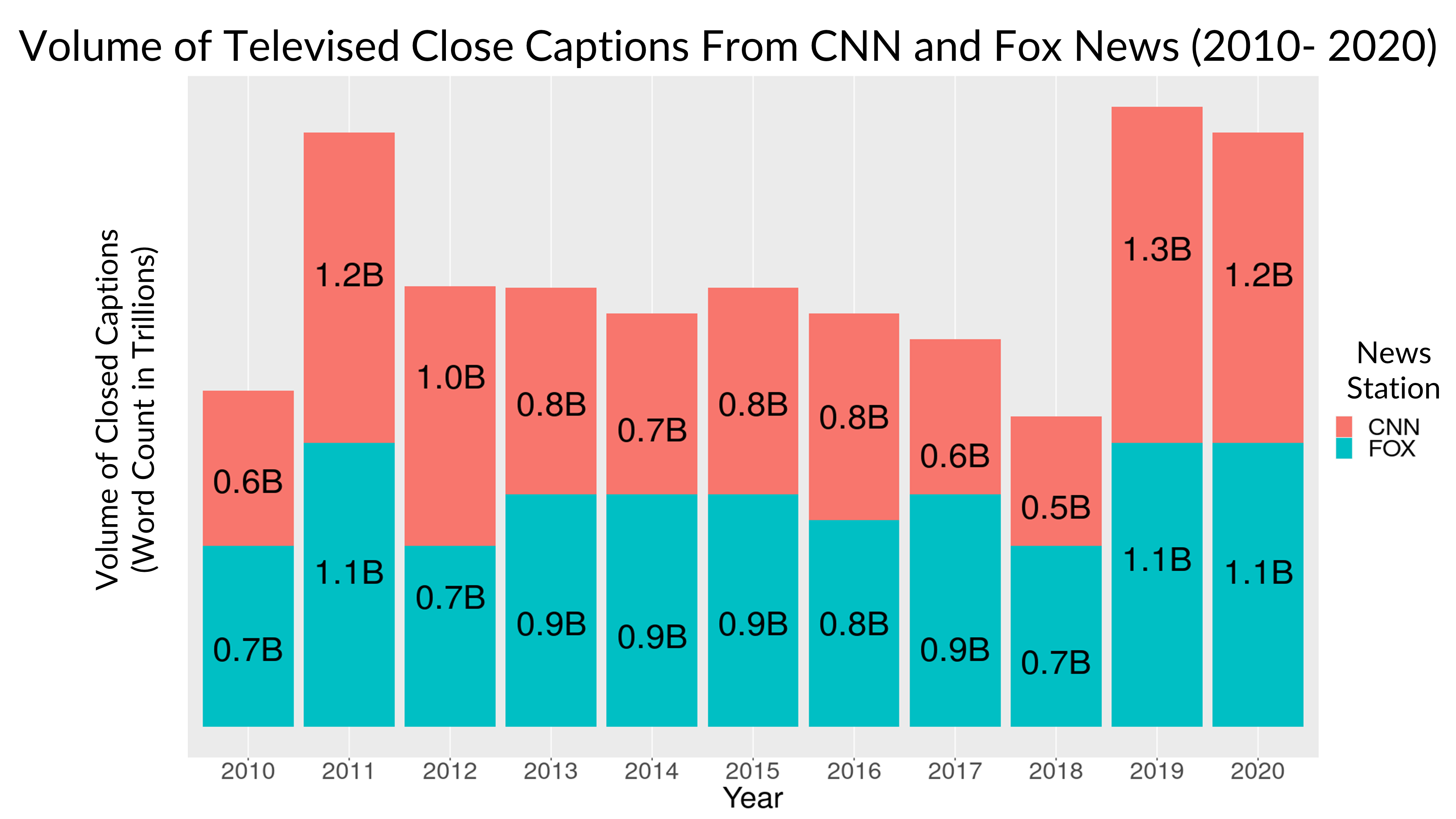} 
\caption{Volume of televised closed captions (word count) from CNN and Fox News stations from 2010 to 2020.}
\label{fig1}
\end{figure}

Our data consists of transcribed closed captions from news shows that were broadcast 24-7 by CNN and Fox News from January 1, 2010 to December 31, 2020, which was provided by the Internet Archive and the Stanford Cable TV News Analyzer. We computationally extracted closed captions pertaining to each televised segment from a total of $181K$ SRT (SubRip Subtitle) files  --- consisting of $23.8$ million speaker turns or $297$ billion words spoken across the decade  (Figure \ref{fig1}). To focus our analysis on how socially important yet politically controversial topics were discussed between the two major news networks, we narrowed down our sample of speaker turns to those that contained keywords associated with six politically contentious topics: \textit{racism}, \textit{Black Lives Matter}, \textit{police}, \textit{immigration}, \textit{climate change}, and\textit{ health care} (Table \ref{table1}). We extracted all sentences containing one of the nine topical keywords, giving us a final corpus of $2.1$ million speaker turns or $19.4$ billion words. All closed captions pertaining to commercials were removed using a matching search algorithm. \cite{tuba_information_2020}.
\subsection{Method and Analysis}
\begin{figure*}[htb]
\centering
\includegraphics[width=1.85\columnwidth]{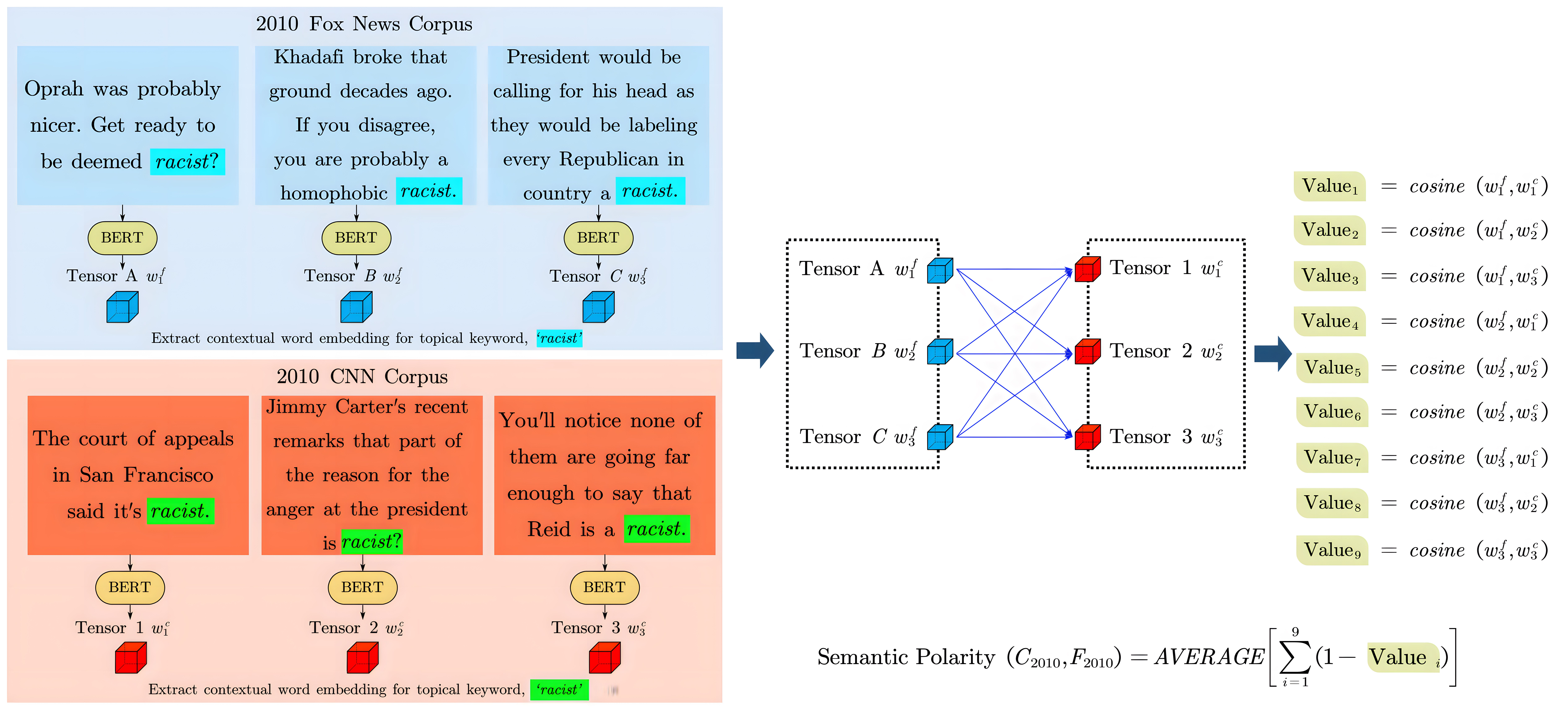}
\caption{Methodological framework of deriving semantic polarization (how CNN \& Fox News use the keyword \textit{racist} in 2010).} 
\label{fig2}
\end{figure*}
\begin{figure*}[!htb]
\centering
\includegraphics[width=1.55\columnwidth]{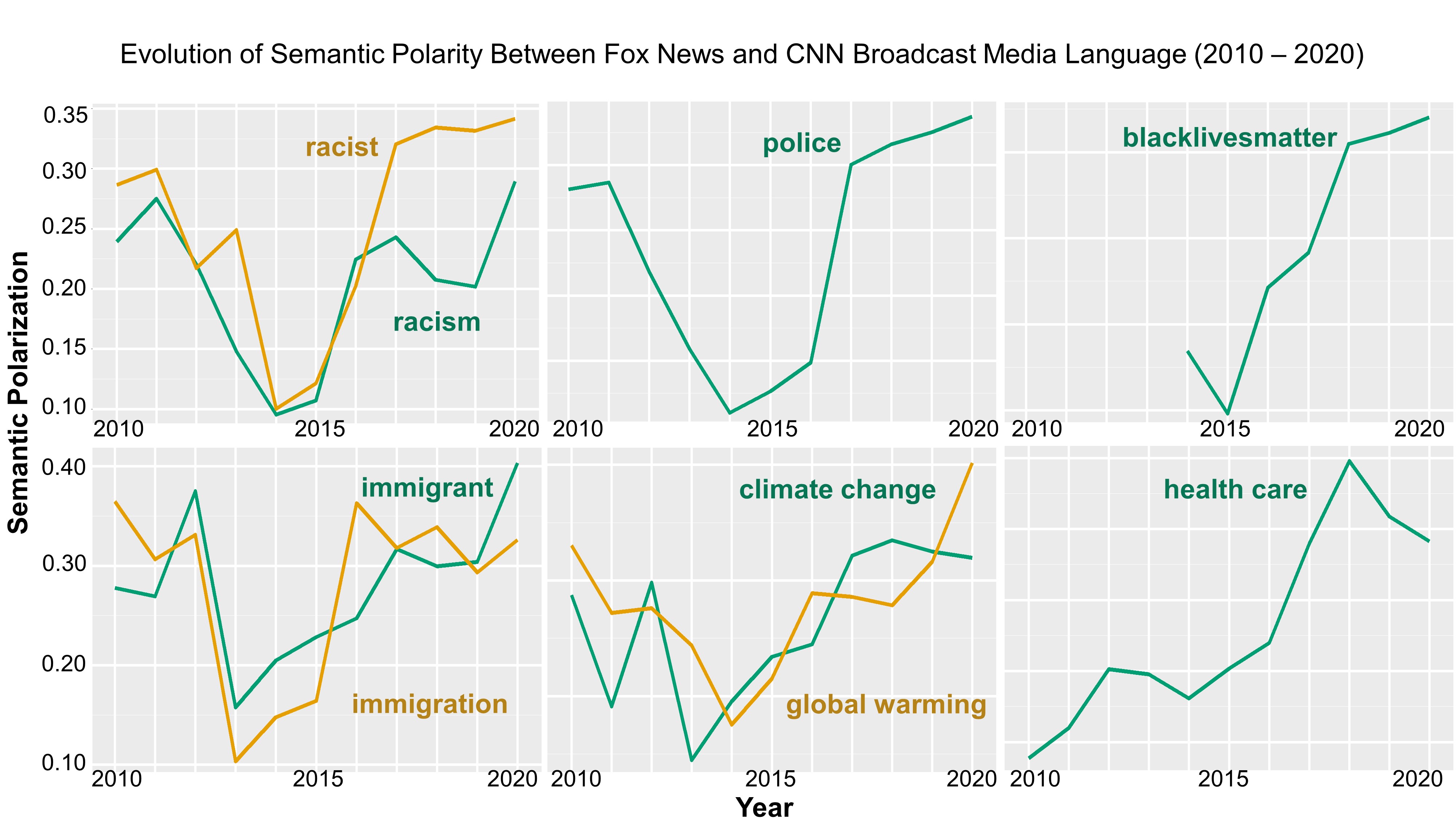}
\caption{Evolution of semantic polarization in how CNN and Fox News use topically contentious keywords from 2010 to 2020.}
\label{trends}
\end{figure*}
With more recent advances in deep learning, researchers have leveraged the ability of transformer-based language models to generate contextualized embeddings to detect the semantic change of words across time \cite{kutuzov_diachronic_2018}. Similarly, we leverage contextual word representations from BERT (Bidirectional Encoder Representations from Transformers) \cite{devlin_bert_2019} to calculate the semantic polarity in how CNN and Fox News use topical keywords over time. 

First, we use BERT to learn a set of embeddings for each  keyword from the hidden layers of the model for every year from 2010 to 2020, to quantify how much the embeddings for a given topical keyword from CNN semantically differ from those of Fox across an 11-year period. We extracted all contextual word representations from our corpus of $2.1$ million speaker turns. Then, for every occurrence of the $9$ topical keywords in the data, we calculated the semantic distance between each contextual word representation from CNN and Fox using cosine values. The use of cosine values of BERT word embeddings is standard in machine translation and other NLP domains for computing contextual similarities in how tokens are used in sentences \cite{zhang_bertscore_2020}. Our calculation of semantic polarization between how CNN vs. Fox News uses a topical keyword for a given year is as follows.

Given the set of word embeddings for a keyword in a given year $t$ from CNN:
\begin{equation}
C_{t}=\left\{w^c_{1}, w^c_{2}, w^c_{3}, \ldots\right\}
\label{eq1}
\end{equation}and the set of word embeddings for a keyword in a given year $t$ from Fox News:
\begin{equation}
F_{t}=\left\{w^f_{1}, w^f_{2}, w^f_{3}, \ldots\right\}
\label{eq2}
\end{equation}
we define the semantic polarity (SP) between CNN and Fox News as:
\begin{equation}
\operatorname{SP}\left(C_{t}, F_{t}\right)=\frac{\sum_{w^c_{i} \in C_{t}, w^f_{j} \in F_{t}} 1 - \cos \left(w^c_{i}, w^f_{j}\right)}{n_{1} n_{2}} \label{con:inventoryflow}
\end{equation}
where:
\begin{conditions*}
C_t & :&set of word embeddings from CNN \\
F_t & :&set of word embeddings from Fox News \\
w^c_i &:&word embedding of a keyword from CNN \\
w^f_j &:&word embedding of a keyword from Fox News \\
n_1 &:&number of word embeddings from CNN \\
n_2 &:&number of word embeddings from Fox News
\end{conditions*}
In Figure \ref{fig2}, we visualize our methodological framework to illustrate how we derive the semantic polarization (SP) score in how CNN vs. Fox News uses the keyword, “racist” in 2010. First, we extract all contextual word embeddings (each of which is a $768 \times 1$ tensor) associated with every occurrence of the word “racist” in our corpus of speaker turns from Fox and CNN in 2010. Next, we calculate the cosine distance between the tensors associated with the word representation “racist” in the CNN corpus and those from the Fox News corpus. Finally, we use Equation \ref{con:inventoryflow} to calculate the average of all cosine distance values to obtain the semantic polarization score or the semantic distance between how Fox News and CNN uses the word ``racist" in 2010.
\subsection{Results}
Between 2010 and 2020, semantic polarity between America’s two largest TV news stations increased with a significant upward trend starting in 2015 and 2016 (Figure \ref{trends}). Over the past decade, CNN and Fox News became increasingly semantically divergent in how they used the topical keywords ``\textit{racist}", ``\textit{racism}", ``\textit{police}", ``\textit{blackbivesmatter}", ``\textit{immigrant}", ``\textit{immigration}", ``\textit{climate change}",``\textit{global warming}", and ``\textit{health care}" in their broadcasting language. Though there is a general dip in semantic polarity around 2013-14 across all keywords with the exception of “\textit{health care}”, semantic polarity between the two stations starts to sharply increase every year on average by $112\%$ across all keywords starting in 2016. These results corroborate findings from a recent study analyzing the airtime of partisan actors on CNN and Fox, which shows that the two stations became more polarized in terms of visibility bias particularly following the 2016 election \cite{kim2022measuring}. \newline
\indent Further, the highest peaks generally occur in 2020 for the terms “\textit{racism}”, “\textit{racist}”, “\textit{blacklivesmatter}”, “\textit{police}”, along with “\textit{immigrant}”, and “\textit{global warming}” (Table \ref{table2}). Specifically, semantic polarity is highest for the keyword “\textit{police}” in 2020, denoting that across the nine keywords, CNN and Fox were most polarized in how they semantically used the word “\textit{police}” in their news programs in 2020.
\raggedbottom
\begin{table}[!htb]
\centering
\resizebox{.84\columnwidth}{!}{
\begin{tabular}{|l|c|c|}
\hline Keywords & Min SP & Max SP\\
\hline racism            &0.095(2014)   &0.289\textbf{(2020)}\\
\hline racist             &{0.100 (2014) }  &0.341\textbf{(2020)}\\
\hline {blacklivesmatter}              &0.148(2015)   &0.320\textbf{(2020)}\\
\hline {police}              &0.119(2014)    &0.574\textbf{(2020)}\\
\hline immigration                  &0.103(2013)            &0.362(2016)\\
\hline {immigrant}           &{0.157(2013)}            &0.383\textbf{(2020)}\\
\hline climate change               &0.144(2013)            &0.334(2018)\\
\hline {global warming}      &0.175(2014)            &0.401\textbf{(2020)}\\
\hline {health care}                  &0.076(2010)            &0.496(2018)\\
\hline
\end{tabular} }
\caption{Range of 2020 semantic polarity scores.}
\label{table2}
\end{table}
\section{Study 2: Words that Characterize Semantic Polarity between Fox News \& CNN in 2020}
\subsection{Method and Analysis}
Semantic polarity between Fox and CNN is predominantly high in 2020, echoing the series of highly publicized events in the media (e.g., COVID-19, George Floyd, and BLM demonstrations) that reified a widening ideological gap among the American public on a number of issues. Hence, in Study 2, we aim to go beyond providing a mere quantification of polarization by examining how the language surrounding the use of contentious keywords by CNN and Fox contributes to the 2020 peak in semantic polarity. In other words, \textbf{\textbf{how does CNN and Fox News \textit{contextually} differ in their use of topical keywords on their news programs in 2020?}} To answer this question, we identified attributive words or contextual tokens that are most predictive of whether a speaker turn containing a topical keyword was from CNN or Fox in 2020. This task entailed a two-step process. First, we trained a BERT-based classifier using the 2020 corpus of closed captions from both stations to predict whether a speaker turn was from CNN or Fox. Then, we used a model interpretation technique called Integrated Gradients \cite{sundararajan_axiomatic_2017} to identify which words (except for the topical keyword) in each speaker turn contributed most to the classification decision.
\subsubsection{Classifying 2020 Speaker Turns.}
To identify whether a speaker turn was broadcast from Fox or CNN in 2020, we built six BERT-based classifiers pertaining to each topic. First, we merged the 2020 corpus of speaker turns, each containing one of the nine topical keywords by topic (e.g., speaker turns containing “racism” or “racist” were merged into one topical corpus).\newline
\indent For each of the six topical corpora, we trained BERT (\texttt{BERT-Base-Uncased}) by masking $15\%$ of the words in each speaker turn and trained the model to fill in the masked words using a standard cross-entropy loss function. We then fine-tuned the adapted model to predict whether a speaker turn was from CNN or Fox News. For this step, we split the data into $80\%$ training, $10\%$ validation, and $10\%$ test sets. We set batch sizes to $16$, training epochs to 6 with $3$-fold cross-validation, and used AdamW for optimization with a learning rate of $2\mathrm{e}{-05}$.  We used PyTorch for all model implementations.
\subsubsection{Model Interpretation.}
To identify the contextual language or the attributive words that contributed most to the 2020 semantic polarity between CNN and Fox News, we used Integrated Gradients (IG) to interpret the classification results from our BERT-classifier in the previous step. IG is an attribution method that allows us to understand how much an input feature contributes to the output result of a deep neural network \cite{sundararajan_axiomatic_2017}. This post-hoc interpretability technique combines the input's gradients by interpolating in short increments along a straight line between a baseline $x$ (usually a vector with all zeros) and the input $x'$. Formally, if $F: R_n \rightarrow [0, 1]$ represents BERT, then the integrated gradient (IG) of the $i^{th}$ dimension of $x$ is:
\begin{equation}
I G_{i}(x)=\left(x_{i}-x_{i}^{\prime}\right) \times \int_{\alpha=0}^{1} \frac{\partial F\left(x^{\prime}+\alpha\left(x-x^{\prime}\right)\right)}{\partial x_{i}} d \alpha
\label{ig}
\end{equation}
\indent Large pre-trained transformer-based language models, such as BERT may have strong predictive powers, but their multiple nonlinearities prevent users from directly attributing model outputs to various input features. To overcome this lack of interpretive insight, we leverage IG to identify words that are most attributable to each predicted class. Thus, we identify which contextual words within a speaker turn (input features of a classifier) contribute most to, or are most predictive of the output decision, namely, whether the speaker turn is from CNN or Fox News in 2020.
\subsection{Results}
Our BERT-classifier distinguishes 2020 speaker turns from CNN and Fox News with a mean accuracy of $75.3\%$ across all six topics (Table \ref{table3}). Model performance is highest for speaker turns topically focused on the police, corresponding to Study 1 findings demonstrating highest SP scores for ``police" across all keywords in 2020.
\begin{table}[!ht]
\centering
\resizebox{0.94\columnwidth}{!}{
\begin{tabular}{|c|c|c|c|c|}
\hline Topic & Acc. & F1 & Prec. & Recall\\
\hline Racism           &0.710&0.709&0.707&0.712\\
\hline BLM              &0.724&0.726&0.710&0.743\\
\hline Police           &0.826&0.829&0.815&0.845\\
\hline Immigration      &0.773&0.780&0.762&0.800\\
\hline Climate change   &0.737&0.710&0.750&0.675\\
\hline Health care      &0.750&0.752&0.748&0.757\\
\hline
\end{tabular} }
\caption{Model performance for station classification.}
\label{table3}
\end{table}

In Table \ref{table4}, we show the top 10 words most predictive of how CNN and Fox News discuss the six topics on their news programs in 2020. To avoid noise caused by low-frequency words, we limited our token selection to words with a frequency above the $95$th percentile. We categorized the tokens with the highest and lowest attribution scores ranging from positive (attributed to CNN) to negative (attributed to Fox News) values for each topic. As shown, across all topics, none of the top tokens overlap between the two stations, indicating the drastically distinct contexts in which the two channels discuss key topics in 2020. For example, words like “\textit{illegal}”, “\textit{enforcement}”, and “\textit{order}”  contextualize immigration discussions in a rather legalistic manner on Fox News whereas, “\textit{parents}”, “\textit{family}”, “\textit{children}”, “\textit{daughter}”, and “\textit{communities}” - humanizing words connotative of family relations best predict how CNN discusses the same topic. Both stations may be discussing the same topic with identical keywords, but the contextual language surrounding their discussion is strikingly different. Such divergence sheds light on the underlying differences in the linguistic signature of televised media language that help explain the 2020 peak in semantic polarity between the two stations.
\section{Study 3: How Semantic Polarization in Broadcast Media Language Forecasts Semantic Polarity in Social Media Discourse }
Study 1 shows that semantic polarization in broadcast media language has been increasing in recent years and overall peaking closer to 2020 during which the two major stations  use starkly contrasting language in their discussion of identical words (Study 2). Such partisan divide in televised news language is similarly observed online where public discourse is becoming increasingly fraught with political polarization \cite{chinn_politicization_2020}. To what degree is broadcast media one of the causes of online polarization in today’s agora of public discourse on social media? \textbf{Do the temporal changes in semantic polarity in broadcast media significantly predict ensuing temporal changes in the semantic polarization across social media discourse?}

To answer this, we used Granger causality to test whether semantic polarity between CNN and Fox News broadcast media language over the last decade Granger-causes the semantic polarity between Twitter users who follow and mention @CNN or @FoxNews in their tweets that contain one of the nine keywords (``\textit{racist}", ``\textit{racism}", ``\textit{police}", ``\textit{blackbivesmatter}", ``\textit{immigrant}", ``\textit{immigration}", ``\textit{climate change}",``\textit{global warming}", and ``\textit{health care}") over the same time period.
\subsection{Twitter Dataset}
For our social media corpus, we collected all tweets between 2010 and 2020 based on the following criteria: (1) written by users who follow both @CNN and @FoxNews; (2) mention or are replying to either @CNN or @FoxNews; (3) contain one of the nine keywords.
Our final corpus amounted to $131,627$ tweets spanning over a decade. The volume of tweets and mean word count per tweet are shown in Table \ref{table5}.
\begin{table*}[!ht]
\centering
 \resizebox{2.11\columnwidth}{!}{
\begin{tabular}{|c|c|c|c|c|c|c|c|c|c|c|c|}
\hline \multicolumn{4}{|c|}{Racism} & \multicolumn{4}{c|}{Black Lives Matter} & \multicolumn{4}{c|}{Climate Change}\\
\hline \multicolumn{2}{|c|}{CNN} & \multicolumn{2}{c|}{Fox News} & \multicolumn{2}{c|}{CNN} & \multicolumn{2}{c|}{Fox News} & \multicolumn{2}{c|}{CNN} & \multicolumn{2}{c|}{Fox News} \\
\hline Token & Attr. & Token & Attr. & Token & Attr. &Token & Attr. & Token & Attr. &Token & Attr.\\
\hline president&0.457&called&-0.482&confederate&0.717&party&-0.505 & human&0.483&gun&-0.440\\
\hline whitehouse&0.348&anti&-0.442&people&0.350&video&-0.475 & science&0.435&left&-0.419\\
\hline senator&0.224&crime&-0.413&we&0.285&shooting&-0.454 & scientists&0.289&democrats&-0.385\\
\hline there&0.155&left&-0.398&symbol&0.281&mayor&-0.440 & states&0.259&democratic&-0.369\\
\hline protests&0.153&accused&-0.380&saw&0.261&anti&-0.400 & today&0.244&agenda&-0.352\\
\hline comments&0.146&school&-0.376&thousands&0.255&weekend&-0.394 & climate&0.236&party&-0.349\\
\hline he&0.141&criminal&-0.330&came&0.249&night&-0.351 & california&0.207&progressive&-0.324\\
\hline conversation&0.128&democratic&-0.325&players&0.247&democratic&-0.349 & believe&0.207&right&-0.276\\
\hline brutality&0.111&government&-0.319&know&0.240&police&-0.340 & issues&0.203&voters&-0.264\\
\hline ago&0.102 &bad&-0.302&they&0.240&democrats&-0.335 & important&0.199&Obama&-0.264\\

\hline \multicolumn{4}{|c|}{Police} & \multicolumn{4}{c|}{Immigration} & \multicolumn{4}{c|}{Health Care}\\
\hline \multicolumn{2}{|c|}{CNN} & \multicolumn{2}{c|}{Fox News} & \multicolumn{2}{c|}{CNN} & \multicolumn{2}{c|}{Fox News} & \multicolumn{2}{c|}{CNN} & \multicolumn{2}{c|}{Fox News} \\
\hline Token & Attr. & Token & Attr. & Token & Attr. &Token & Attr. & Token & Attr. & Token & Attr. \\
\hline different&0.414&seattle&-0.641&racist&0.370&illegal&-0.787 & vaccines&0.735&illegal&-0.647\\
\hline they&0.351&city&-0.537&Trump&0.298&laws&-0.422 & vaccine&0.525&businesses&-0.568\\
\hline certainly&0.346&portland&-0.512&parents&0.254&enforcement&-0.367 & va&0.453&free&-0.477\\
\hline racial&0.295&minnesota&-0.489&obama&0.252&police&-0.308&masks&0.386&business&-0.459\\
\hline actually&0.285&chicago&-0.446&pan&0.212&legal&-0.303&deaths&0.336&open&-0.361\\
\hline understand&0.285&left&-0.414&family&0.210&order&-0.288&virus&0.323&pay&-0.352\\
\hline know&0.285&street&-0.406&vice&0.208&million&-0.278&faucci&0.295&economy&-0.338\\
\hline shows&0.278&riots&-0.401&daughter&0.186&protect&-0.261&climate&0.283&jobs&-0.328\\
\hline hands&0.269&democratic&-0.397&children&0.167&wants&-0.222&weeks&0.273&safe&-0.318\\
\hline we&0.267&crime&-0.395&communities&0.167&law&-0.213&Trump&0.259&city&-0.311\\
\hline
\end{tabular} 
 }
\caption{Top 10 tokens with the highest (attributed to CNN) and lowest (attributed to Fox News) attribution scores as a proxy for understanding words that linguistically characterize the 2020 semantic polarity between Fox News and CNN across six topics.}
\label{table4}
\end{table*}
\subsection{Hypothesis Testing with Granger Causality}
To test whether temporal patterns in semantic polarity in TV news language significantly predict semantic polarity trends across Twitter users, we conducted a Granger causality analysis. Granger causality is a statistical  test to determine whether a time series $X$ is meaningful in forecasting another time series $Y$ \cite{granger_testing_1980}. 
Simply, for two aligned time series $X$ and $Y$, it can be said that $X$ Granger-causes $Y$ if past values $ X_{t-l} \in X $ lead to better predictions of the current $Y_t \in Y$ than do the past values $Y_{t-l} \in Y$ alone, where $t$ is the time point and $l$ is the lag time or the time interval unit in which changes in $X$ are observed in $Y$. Using this logic, we tested the following hypotheses:
\begin{itemize}
    \item[] \textit{\textbf{H1.}}\;Semantic polarization on TV news significantly Granger-causes semantic polarization on Twitter.
    \item[] \textit{\textbf{H2.}}\;Semantic polarization on Twitter significantly Granger-causes semantic polarization on TV.
\end{itemize}
\raggedbottom
\begin{table}[!ht]
\centering
\resizebox{1.0\columnwidth}{!}{
\begin{tabular}{|l|c|c|c|c|}
\hline \multicolumn{1}{|c|}{} & \multicolumn{2}{c}{Tweet Volume} & \multicolumn{2}{|c|}{\makecell[c]{Avg. Word Count \\(per Tweet)}}\\
\hline \multicolumn{1}{|l|}{Keyword}&FoxNews&CNN&FoxNews&CNN\\
\hline racism&8,787&8,944&24&22\\
\hline racist&7,887&8,795&27&22\\
\hline blacklivesmatter&4,863&4,528&21&23\\
\hline police&8,731&8,670&24&22\\
\hline immigration&8,163&7,883&19&21\\
\hline immigrant&8,239&7,721&20&28\\
\hline climate change&7,536&7,140&21&22\\
\hline global warming&5,042&5,140&18&17\\
\hline health care &7,612&6,943&20&19\\
\hline
\end{tabular} }
\caption{Volume of tweets and mean word account.}
\label{table5}
\end{table}
\subsection{Method and Analysis}
First, we computed the monthly semantic polarity (SP) scores spanning from 2010 to 2020 for a total of $132$ time points (12 months $\times$ 11 years) for each topical keyword for both the televised closed caption and Twitter data. We then built $18$ time series using the monthly SP values for each of the $9$ keywords across $11$ years derived from the closed captions and the tweets. We then paired each of the TV news language time series with the Twitter time series by keyword, giving us a total of 9 causality tests per hypothesis. As a prerequisite for Granger causality testing, we conducted the Augmented Dickey-Fuller (ADF) test \cite{cheung_lag_1995} to ensure that the value of the time series was not merely a function of time (see Appendix). We used the serial difference method \cite{cheung_lag_1995} to obtain stationarity for $\Delta TwS_{racism}$ and $\Delta TvS_{BLM}$, which were the only two time series with an ADF test value greater than the 5\% threshold. In total, we conducted 18 time series analyses to test whether there was a  Granger-causal relationship between the monthly SP values of broadcast media language ($SP_{tv}$) and that of Twitter ($SP_{tw}$) by each of the 9 keywords to determine the extent to which semantic polarization on TV drives semantic polarization on Twitter (\textit{H1}) and vice versa (\textit{H2}). For both hypotheses, we used a time-lag ranging from 1-12 months.
\raggedbottom
\begin{table}[!b]
    \centering
    \resizebox{0.95\columnwidth}{!}{
    \begin{tabular}{|c|c|c|c|c|}
\hline {Keyword} & \makecell[c]{Hypos} & \makecell[c]{Lag} & {$F$-value} & $p$-value \\
\hline
\multirow{2}{*}{\makecell[c]{racism}} & H1 & 2 & 1.70      &  \textbf{0.0386}\\ \cline{2-5} 
                        & H2 & 2 & 0.19      & 0.4526 \\ \hline
\multirow{2}{*}{racist} & H1  &2 & 3.17  & \textbf{0.0101}  \\ \cline{2-5}                   
                       & H2 & 5 &  0.36 & 0.8730          \\ \hline
     \multirow{2}{*}{blacklivesmatter}                & H1             &3 & 2.90 & \textbf{0.0376}   \\ \cline{2-5}                                & H2 & 3 & 3.25  & \textbf{0.0241} \\ \hline
     \multirow{2}{*}{police}                       & H1                    &3 & 0.75 & 0.5189            \\ \cline{2-5}                         & H2 & 5 & 0.50  & 0.7747          \\ \hline
     \multirow{2}{*}{immigration}             & H1          &3 & 2.13 & \textbf{0.0333}   \\ \cline{2-5}                                 & H2 & 1 & 1.11  & 0.2929           \\ \hline
     \multirow{2}{*}{immigrant}               & H1            &3 & 2.19 & \textbf{0.0325}   \\ \cline{2-5}                               & H2 &1 & 1.11  & 0.2929           \\ \hline
    \multirow{2}{*}{climate change}       & H1    &7 & 2.00 & 0.0502            \\ \cline{2-5}                                        & H2 &3 & 3.18  & \textbf{0.0262}  \\ \hline
    \multirow{2}{*}{global warming}      & H1  &  8 & 0.96  & 0.4691         \\ \cline{2-5}                                        & H2 & 4 & 2.15 & \textbf{0.0213}  \\ \hline
     \multirow{2}{*}{heath care}               & H1            &4 & 0.58 & 0.6738            \\ \cline{2-5}                               & H2 &3 & 0.72  & 0.5377           \\ \cline{2-5}
    \hline
    \end{tabular} }
    \caption{H1 and H2 results using Granger causality tests. H1:  $SP_{tv}$ $\not\to\mathcal{}$  $SP_{tw}$; H2: $SP_{tw}$ $\not\to\mathcal{}$  $SP_{tv}$.}
    \label{table granger}
    \end{table}
\subsection{Results}
We observe that changes in semantic polarity patterns on Twitter can be significantly predicted by semantic polarity trends in broadcast media language and vice-versa. Table \ref{table granger}, shows Granger causality results for hypotheses 1 and 2 with corresponding significant lag lengths ($p < 0.05$).
\subsubsection{Hypothesis 1.} As shown in Table \ref{table granger}, semantic polarity trends in broadcast media’s discussion on racism (keywords: “\textbf{racism}”, “\textbf{racist}”) and immigration (keywords:  “\textbf{immigration}” and “\textbf{immigrant}”) significantly forecasts semantic polarity shifts in how people discuss these topics on Twitter with a time lag of 2 and 3 months, respectively. In other words, it takes about 2 months for the influence of semantic polarization in broadcast media language to manifest across how Twitter users semantically diverge in their discussion on racism, and about 3 months in how they talk about keywords linked to immigration.

\subsubsection{Hypothesis 2.}  To test whether the directionality of influence is not only such that semantic polarization in televised news shapes polarity across Twitter conversations, but also vice-versa, we tested \textit{H2}. Our results show that for “\textbf{climate change}” and “\textbf{global warming}”, semantic polarization across how people discuss these keywords on Twitter Granger-causes how CNN and Fox News semantically diverge in their use of these words on their shows with a lag of 3 months. This may be due to the fact that discussions on climate change and global warming have become increasingly political and less scientific on Twitter \cite{jang_polarized_2015, chinn_politicization_2020}, therefore raising online prominence of the topic through controversial political sound bites that may in return, draw attention from mainstream media outlets as a source for news reporting.

\subsubsection{Bidirectional Causality.}  For Black Lives Matter, there is bidirectional causality, meaning semantic polarity in how the keyword ``\textbf{blacklivesmatter}" is discussed on televised news programs shapes semantic polarity trends in how users talk about the topic and vice-versa. This may be due to the fact that BLM is an online social movement, largely powered by discursive networks on social media. Hence, while users may draw upon the narratives generated by traditional media outlets in their Twitter discussions, journalists and broadcast media companies too, may be attuned to how online users discuss BLM issues on social media.
\subsection{Words That Characterize the Relational Trends in Semantic Polarization  Between Broadcast Media and Twitter Discourse}
To understand the manner in which language plays a role in how semantic polarity trends in broadcast media forecasts polarization patterns on Twitter, for topical keywords with significant Granger-causality results, we identified contextual tokens that are most characteristic of how each topical keyword is discussed on TV vs. Twitter separated by the corresponding lag lengths based on results shown in Table \ref{table granger}. 

Hence, given the 132 time points (12 months $\times$ 11 years) represented by $t_{ij}$, where $i$ represents the month $( i \in [ 1, 12])$, $j$ is the year $( j \in [1, 11])$, and $l$ is the value of the lag length in months corresponding to significant Granger coefficients $(l \in [1, 8])$, we first re-organized our TV news data with $t_{i^{\prime} j}$ time points, where ${i^{\prime}}$ represents the value of the maximum month minus $l$ with the range: $\left(i^{\prime} \in[1,12-l]\right)$ and our Twitter data with $t_{i^{\prime\prime} j}$ time points, where ${i^{\prime\prime}}$ represents the value of the minimum month plus $l$ with the range: $\left(i^{\prime\prime} \in[1+l,12]\right)$. For topical keywords where semantic polarity in Twitter Granger-causes semantic polarity trends in broadcast media language, we interchange the values of $i^{\prime}$ and $i^{\prime\prime}$. Next, following a similar approach to Study 2, we separately fed the TV news and Twitter corpora into a BERT-classifier and used Integrated Gradients to identify the top 10 most predictive tokens for each topical keyword. 

Results are shown in Tables \ref{table b-1} - \ref{table b-4} (Appendix) where we demonstrate the top 10 contextual tokens that are most attributive of how each topical keyword is used either by a speaker from CNN or Fox News stations or by a user responding to @CNN or @FoxNews in their tweet during time periods separated by monthly lag lengths corresponding to significant Granger-causal relationships. 

As shown by the bolded tokens,  a fair proportion of the top 10 words most characteristic of how televised news programs (CNN vs. Fox News) and Twitter users (replying to @CNN vs. @FoxNews) discuss topical keywords overlap quite a bit, suggesting that these tokens are semantically indicative of how linguistic polarity in one media manifests in another over time. For example, six of the top 10 tokens predictive of how CNN broadcast news contextually uses the keyword ``racism" are identical to that of how users on Twitter, replying to @CNN, use ``racism" in their tweets: ``believe", ``today", ``president", ``Trump", ``campaign", ``history", ``issue". 

For ``blacklivesmatter" where the Granger-causality relationship is bidirectional, words most predictive of semantic polarity in broadcast news that drive polarized discourse on Twitter are different from the tokens on Twitter that drive polarity in televised news language. For example, the set of tokens (``violence", ``matter", ``white", ``support", ``blacks") most predictive of how TV news programs on Fox News talk about BLM that appear three months later in tweets replying to @FoxNews, are strikingly different from the set of words (``blacks", ``cities", ``democrats", ``lives", ``moral", ``organization", and ``shoot") that best characterize the tweets replying to @FoxNews, which appear three months later on the Fox News channel.
\section{Discussion}
The rising levels of semantic polarity between the two major broadcast news organizations, as demonstrated in our work may render people's ability to reach across partisan divides and to perceive and solve issues democratically more difficult. The way CNN and Fox News discuss identical keywords on their programs is remarkably distinct in 2020 where semantic polarity between the two stations reaches its peak over an 11-year period, corroborating the widening partisan media gap highlighted by recent scholarship. Framing Theory argues that even subtle changes in the wording around how an issue is described by the media can significantly influence how audiences interpret the issue \cite{scheufele2000agenda, entman2003cascading}. Yet, our findings show that the contextual language driving semantic polarization in broadcast media is not nuanced at all. The surrounding words that characterize how each station discusses topically important keywords are drastically different, to the extent that identical words seem to reflect different meanings altogether. This is a crucial point of consideration, as linguistic cues in the media can play a powerful role in selectively reifying certain aspects of the perceived reality of an issue while downplaying others. Our findings suggest that TV news language does not only shape how online audiences perceive issues, but also how they talk about them. 

In linguistics, collocation is the property of language where two or more words appear in each other’s company with greater than random chance \cite{hoey_lexical_2005}, such as “illegal” and “immigrants” in Fox News and “climate change” and “science” from CNN, as shown in our findings. Repeated encounters with collocated words drive what linguists call lexical priming \cite{hoey_lexical_2005}, in which a priming word (e.g., "blacklivesmatter") provokes a particular target word (“protests” for CNN and “violence” for Fox) more strongly and quickly than disparate words encountered more rarely. In this vein, online audiences who consume very different perceived realities from the media may be lexically primed through repeated exposure to collocated words that frame identical issues in acutely contrasting contexts. This may help theorize why TV news language can shape how online audiences engage in public discourse as demonstrated in this work. Semantic polarization in televised media not only forecasts semantic polarity trends on Twitter, but the \textit{words} that characterize broadcast media polarization re-appear across Twitter discussions, separated by significant lag months. Our results demonstrate that the language characterizing opposing media narratives around topically similar news events on TV can be linguistically mimicked in how social media users are polarized in their discourse around important social issues.

\textit{\textbf{Limitations}. }Differences in cosine values between contextual word representations stem not only from semantic, but also from positional differences of words in sentences. Although word embeddings generally embody contextual nuances, we acknowledge that our work predominantly considers semantic rather than syntactic aspects, and that to some extent our calculation of semantic polarity could embody syntactic differences in how keywords are discussed. Furthermore, our findings pertain to American political contexts, which might not be generalizable to foreign public discourse and media language. For future work, we aim to apply our model to relevant textual corpora in other languages by incorporating cross-lingual transfer learning methods to better understand the generalizability of our model by comparing distributional differences with cross-lingual data. 

\section{Ethics Statement}
We aim to strictly adhere to the AAAI Code of Ethics and Conduct by ensuring that no personal information of Twitter users was collected nor compromised throughout our project. All data in this work are securely stored on servers only accessible by the authors. Our semantic polarization framework is made publicly available on the authors' GitHub page and can be applied on a variety of textual corpora with diachronic and contrastive properties.

\appendix
\section{Appendix}
\begin{table}[!ht]
\centering
\resizebox{0.9\columnwidth}{!} {
\begin{tabular}{|c|c|c|c|c|}
  \hline \makecell[c]{Time\\Series} & ADF & \makecell[c]{Threshold\\(1\%)} & \makecell[c]{Threshold\\(5\%)} & Conclusion \\
  \hline $TwS_1$ & -8.52 & -3.48 & -2.88 & stat \\
  \hline $TwS_2$ & -1.86 & -3.45 & -2.88 & {non-stat} \\
  \hline $\Delta TwS_2$ & -5.28 & -2.69 & -1.85 & stat \\
  \hline $TwS_3$ & -4.12 & -3.43 & -2.82 & stat \\
  \hline $TwS_4$ & -7.43 & -3.48 & -2.87 & stat \\
  \hline $TwS_5$ & -8.01 & -3.48 & -2.85 & stat \\
  \hline $TwS_6$ & -6.80 & -3.47 & -2.88 & stat\\
  \hline $TwS_7$ & -4.12 & -3.48 & -2.82 & stat\\
  \hline $TwS_8$ & -7.78 & -3.46 & -2.84 & stat\\
  \hline $TwS_9$ & -5.78 & -3.48 & -2.82 & stat\\
      \hline $TvS_1$ & -4.48 & -3.48 & -2.88 & stat \\
      \hline $TvS_2$ & -7.87 & -3.48 & -2.87 & \makecell[c]{stat} \\
      \hline $TvS_3$ & -2.30 & -3.21 & -2.88 & non-stat \\
      \hline $\Delta TvS_3$ & -8.53 & -2.35 & -1.74 & stat \\
      \hline $TvS_4$ & -3.59 & -3.43 & -2.83 & stat \\
      \hline $TvS_5$ & -8.85 & -3.48 & -2.83 & stat \\
      \hline $TvS_6$ & -3.79 & -3.41 & -2.86 & stat\\
      \hline $TwS_7$ & -5.22 & -3.42 & -2.81 & stat\\
      \hline $TwS_8$ & -7.17 & -3.42 & -2.89 & stat\\
      \hline $TwS_9$ & -4.45 & -3.49 & -2.88 & stat\\
      \hline 
    \end{tabular} }
    \caption{ADF test results showing whether the Twitter and TV news time series data are stationary (stat) or non-stationary (non-stat).}
    \label{table adf2}
    \end{table}
\begin{table*}[!ht]
	\centering
	\resizebox{2.15\columnwidth}{!}{
		\begin{tabular}{|c|c|c|c|c|c|c|c|c|c|c|c|c|c|c|c|}
			\hline \multicolumn{8}{|c|}{Keyword: “racism” (lag: 2 months)}           & \multicolumn{8}{|c|}{Keyword: “racist” (lag: 2 months)}                                                                                                                                                                                                                                                                                                                                                                                                                                            \\
			\hline \multicolumn{4}{|c|}{CNN}                                         & \multicolumn{4}{|c|}{FoxNews}                                     & \multicolumn{4}{|c|}{CNN}                       & \multicolumn{4}{|c|}{FoxNews}                                                                                                                                                                                                                                                                                                                                                \\
			\hline \multicolumn{2}{|c|}{TV}                                & \multicolumn{2}{|c|}{Twitter (+2 months)}                         & \multicolumn{2}{|c|}{TV}              & \multicolumn{2}{|c|}{Twitter (+2 months)}  & \multicolumn{2}{|c|}{TV}              & \multicolumn{2}{|c|}{Twitter (+2 months)}  & \multicolumn{2}{|c|}{TV}              & \multicolumn{2}{|c|}{Twitter (+2 months) }                                                                                                                                     \\
			\hline Token                                                             & Attr.                                                             & Token                                           & Attr.                                      & Token                                           & Attr.                                      & Token                                           & Attr.                                      & Token                  & Attr. & Token                  & Attr. & Token                 & Attr.  & Token                 & Attr.  \\
			\hline \textbf{believe}                                                  & 0.25                                                              & \textbf{believe}                                & 0.18                                       & \textbf{right}                                  & -0.30                                      & system                                          & -0.17                                      & systemic               & 0.27  & \textbf{Trump}         & 0.20  & \textbf{crime}        & -0.37  & \textbf{Africa}       & -0.22  \\
			\hline men                                                               & 0.25                                                              & \textbf{issues}                                 & 0.18                                       & \textbf{party}                                  & -0.29                                      & basis                                           & -0.13                                      & \textbf{president}     & 0.25  & terrorism              & 0.19  & states                & -0.36  & \textbf{politics}     & -0.20  \\
			\hline Americans                                                         & 0.24                                                              & \textbf{history}                                & 0.18                                       & police                                          & -0.29                                      & \textbf{party}                                  & -0.12                                      & \textbf{hate}          & 0.22  & children               & 0.19  & \textbf{politics}     & -0.31  & \textbf{crime}        & -0.20  \\
			\hline \textbf{today}                                                    & 0.24                                                              & he                                              & 0.17                                       & marginalized                                    & -0.29                                      & \textbf{media}                                  & -0.12                                      & country                & 0.20  & \textbf{president}     & 0.17  & \textbf{Africa}       & -0.30  & standard              & -0.15  \\
			\hline \textbf{president}                                                & 0.22                                                              & \textbf{presidents}                             & 0.16                                       & \textbf{black}                                  & -0.29                                      & Africa                                          & -0.12                                      & think                  & 0.18  & \textbf{hate}          & 0.14  & black                 & -0.29  & Africans              & -0.14  \\
			\hline \textbf{Trump}                                                    & 0.22                                                              & we                                              & 0.16                                       & \textbf{media}                                  & -0.28                                      & diversity                                       & -0.12                                      & trying                 & 0.17  & \textbf{party}         & 0.13  & Americans             & -0.22  & parents               & -0.13  \\
			\hline understand                                                        & 0.21                                                              & hate                                            & 0.15                                       & Obama                                           & -0.26                                      & guilty                                          & -0.11                                      & \textbf{Trump}         & 0.17  & white                  & 0.12  & right                 & -0.21  & share                 & -0.12  \\
			\hline \textbf{campaign}                                                 & 0.17                                                              & \textbf{today}                                  & 0.14                                       & anti                                            & -0.21                                      & \textbf{right}                                  & -0.10                                      & \textbf{years}         & 0.16  & bias                   & 0.12  & white                 & -0.20  & please                & -0.11  \\
			\hline \textbf{history}                                                  & 0.17                                                              & \textbf{campaign}                               & 0.13                                       & law                                             & -0.21                                      & suicide                                         & -0.11                                      & history                & 0.16  & \textbf{years}         & 0.11  & police                & -0.16  & \textbf{surveillance} & -0.08  \\
			\hline \textbf{issue}                                                    & 0.13                                                              & \textbf{Trump}                                  & 0.11                                       & problem                                         & -0.20                                      & \textbf{black}                                  & -0.09                                      & \textbf{party}         & 0.15  & presents               & 0.10  & \textbf{surveillance} & -0.15  & reforms               & -0.08  \\
			\hline
\end{tabular}}
\caption{Top 10 tokens most predictive of how CNN and Fox News TV stations and Twitter users replying to @CNN and @FoxNews use keywords topically related to \textit{racism}.}
\label{table b-1}
\end{table*}
\begin{table*}[!ht]
	\centering
	\resizebox{2.15\columnwidth}{!}{
		\begin{tabular}{|c|c|c|c|c|c|c|c|c|c|c|c|c|c|c|c|}            
			\hline \multicolumn{8}{|c|}{Keyword: “immigrant” (lag: 3 months)}        & \multicolumn{8}{|c|}{Keyword: “immigration” (lag: 3 months)}                                                                                                                                                                                                                                                                                                                                                                                                                                       \\
			\hline \multicolumn{4}{|c|}{CNN}                                         & \multicolumn{4}{|c|}{FoxNews}                                     & \multicolumn{4}{|c|}{CNN}                       & \multicolumn{4}{|c|}{FoxNews}                                                                                                                                                                                                                                                                                                                                                \\
			\hline \multicolumn{2}{|c|}{TV}                                & \multicolumn{2}{|c|}{Twitter (+3 months) }                        & \multicolumn{2}{|c|}{TV}              & \multicolumn{2}{|c|}{Twitter (+3 months) } & \multicolumn{2}{|c|}{TV}              & \multicolumn{2}{|c|}{Twitter (+3 months) } & \multicolumn{2}{|c|}{TV}              & \multicolumn{2}{|c|}{Twitter (+3 months) }                                                                                                                                     \\
			\hline Token                                                             & Attr.                                                             & Token                                           & Attr.                                      & Token                                           & Attr.                                      & Token                                           & Attr.                                      & Token                  & Attr. & Token                  & Attr. & Token                 & Attr.  & Token                 & Attr.  \\
			\hline tax                                                               & 0.26                                                              & charged                                         & 0.20                                       & bid                                             & -0.26                                     & killer                                          & -0.22                                     & come                   & 0.25 & \textbf{white}         & 0.22 & law                   & -0.26 & \textbf{legally}      & -0.24 \\
			\hline \textbf{united}                                                   & 0.26                                                             & make                                            & 0.20                                      & \textbf{health}                                 & -0.25                                     & \textbf{right}                                  & -0.21                                     & family                 & 0.24 & law                    & 0.21 & people                & -0.24 & scouts                & -0.20 \\
			\hline Trump                                                             & 0.26                                                             & \textbf{united}                                 & 0.19                                      & \textbf{immigration}                            & -0.25                                     & \textbf{immigration}                            & -0.21                                     & \textbf{united}        & 0.23 & \textbf{united}        & 0.19 & \textbf{policies}     & -0.22 & \textbf{policies}     & -0.17 \\
			\hline \textbf{country}                                                  & 0.25                                                             & parents                                         & 0.17                                      & \textbf{right}                                  & -0.24                                     & healthcare                                      & -0.18                                     & \textbf{rights}        & 0.22 & \textbf{Russia}        & 0.18 & \textbf{public}       & -0.20 & international         & -0.17 \\
			\hline care                                                              & 0.24                                                             & family                                          & 0.16                                      & state                                           & -0.20                                     & \textbf{American}                               & -0.16                                     & \textbf{country}       & 0.22 & \textbf{country}       & 0.17 & president             & -0.18 & safety
			                                                                         & -0.16              \\
			\hline know                                                              & 0.24                                                             & temporarily                                     & 0.15                                      & people                                          & -0.19                                     & peaceful                                        & -0.14                                     & \textbf{security}      & 0.21 & visa                   & 0.16 & \textbf{legally}      & -0.17 & \textbf{public}       & -0.15 \\
			\hline think                                                             & 0.21                                                             & \textbf{country}                                & 0.14                                      & \textbf{president}                              & -0.15                                     & \textbf{health}                                 & -0.13                                     & Trump                  & 0.21 & \textbf{security}      & 0.15 & federal               & -0.16 & \textbf{illegal}      & -0.14 \\
			\hline white                                                             & 0.20                                                             & \textbf{immigrants}                             & 0.14                                      & \textbf{American}                               & -0.15                                     & \textbf{president}                              & -0.12                                     & \textbf{white}         & 0.20 & court                  & 0.12 & system                & -0.15 & election              & -0.14 \\
			\hline \textbf{immigrants}                                               & 0.19                                                             & prison                                          & 0.13                                      & law                                             & -0.11                                     & liberal                                         & -0.12                                     & \textbf{Russia}        & 0.19 & police                 & 0.12 & order                 & -0.15 & praises               & -0.11 \\
			\hline lawyer                                                            & 0.18                                                             & obviously                                       & 0.11                                      & new                                             & -0.10                                     & Trump                                           & -0.12                                     & way                    & 0.19 & \textbf{rights}        & 0.10 & \textbf{illegal}      & -0.15 & Trump                 & -0.10 \\
			\hline
\end{tabular}}
\caption{Top 10 tokens most predictive of how CNN and Fox News TV stations and Twitter users replying to @CNN and @FoxNews use keywords topically related to \textit{immigration}.}
\label{table b-2}
\end{table*}
\begin{table*}[!ht]
	\centering
	\resizebox{2.15\columnwidth}{!}{
		\begin{tabular}{|c|c|c|c|c|c|c|c|c|c|c|c|c|c|c|c|}
			\hline \multicolumn{8}{|c|}{Keyword: “climate change” (lag: 3 months)}   & \multicolumn{8}{|c|}{Keyword: “global warming” (lag: 4 months)} \\
			\hline \multicolumn{4}{|c|}{CNN}                                         & \multicolumn{4}{|c|}{FoxNews}                                     & \multicolumn{4}{|c|}{CNN}                       & \multicolumn{4}{|c|}{FoxNews}                                                                                                                                                                                                                                                                                                                                                \\
			\hline \multicolumn{2}{|c|}{TV (+3 months) }                   & \multicolumn{2}{|c|}{Twitter}                                     & \multicolumn{2}{|c|}{TV (+3 months) } & \multicolumn{2}{|c|}{Twitter}              & \multicolumn{2}{|c|}{TV (+4 months) } & \multicolumn{2}{|c|}{Twitter}              & \multicolumn{2}{|c|}{TV (+4 months) } & \multicolumn{2}{|c|}{Twitter}                                                                                                                                                  \\
			\hline Token                                                             & Attr.                                                             & Token                                           & Attr.                                      & Token                                           & Attr.                                      & Token                                           & Attr.                                      & Token                  & Attr. & Token                  & Attr. & Token                 & Attr.  & Token                 & Attr.  \\
			\hline \textbf{science}                                                  & 0.28                                                             & \textbf{years}                                  & 0.24                                      & president                                       & -0.26                                     & \textbf{policy}                                 & -0.23                                     & \textbf{important}     & 0.24 & human                  & 0.23 & \textbf{Obama}        & -0.23 & \textbf{Obama}        & -0.24 \\
			\hline \textbf{years}                                                    & 0.24                                                             & threatened                                      & 0.23                                      & \textbf{right}                                  & -0.25                                     & job                                             & -0.22                                     & \textbf{water}         & 0.23 & drugs                  & 0.23 & future                & -0.23 & hurricanes            & -0.24 \\
			\hline primarily                                                         & 0.24                                                             & \textbf{science}                                & 0.21                                      & \textbf{democrats}                              & -0.22                                     & struggle                                        & -0.22                                     & \textbf{science}       & 0.23 & government             & 0.22 & national              & -0.22 & protect               & -0.23  \\
			\hline time                                                              & 0.24                                                             & national                                        & 0.21                                      & \textbf{economy}                                & -0.22                                     & presidential                                    & -0.21                                     & lives                  & 0.22 & \textbf{water}         & 0.20 & rights                & -0.22 & communities           & -0.22 \\
			\hline people                                                            & 0.23                                                             & geography                                       & 0.21                                      & jobs                                            & -0.22                                     & \textbf{right}                                  & -0.20                                     & \textbf{health}        & 0.22 & \textbf{condition} & 0.19 & \textbf{Trump}        & -0.21 & \textbf{Trump}        & -0.22  \\
			\hline health                                                            & 0.23                                                             & diseases                                        & 0.19                                      & \textbf{policy}                                 & -0.22                                     & violence                                        & -0.20                                     & \textbf{condition} & 0.20 & teenager               & 0.18 & \textbf{gas}          & -0.20 & \textbf{forest}       & -0.21 \\
			\hline \textbf{world}                                                    & 0.23                                                             & \textbf{ocean}                                  & 0.18                                      & care                                            & -0.21                                     & diet                                            & -0.19                                     & world                  & 0.19 & \textbf{health}        & 0.17 & \textbf{economy}      & -0.20 & \textbf{economy}      & -0.17 \\
			\hline think                                                             & 0.22                                                             & \textbf{issues}                                 & 0.13                                      & energy                                          & -0.20                                     & \textbf{economy}                                & -0.19                                     & fossil                 & 0.19 & marketing              & 0.16 & energy                & -0.20 & they                  & -0.17 \\
			\hline \textbf{ocean}                                                    & 0.22                                                             & \textbf{world}                                  & 0.11                                      & \textbf{Trump}                                  & -0.18                                     & \textbf{democrats}                              & -0.19                                     & issue                  & 0.19 & \textbf{science}       & 0.16 & immigration           & -0.19 & USA                   & -0.17 \\
			\hline \textbf{issues}                                                   & 0.21                                                             & predator                                        & 0.10                                      & global                                          & -0.16                                     & \textbf{Trump}                                  & -0.17                                     & environment            & 0.17 & \textbf{important}     & 0.14 & green                 & -0.17 & \textbf{gas}          & -0.15 \\
			\hline
\end{tabular}}
\caption{Top 10 tokens most predictive of how CNN and Fox News TV stations and Twitter users replying to @CNN and @FoxNews use keywords topically related to \textit{climate change}.}
\label{table b-3}
\end{table*}
\begin{table*}[!ht]
	\centering
	\resizebox{2.15\columnwidth}{!}{
		\begin{tabular}{|c|c|c|c|c|c|c|c|c|c|c|c|c|c|c|c|}
			\hline \multicolumn{8}{|c|}{Keyword: “BlackLivesMatter” (lag: 3 months)} & \multicolumn{8}{|c|}{Keyword: “BlackLivesMatter” (lag: 3 months)}                                                                                                                                                                                                                                                                                                                                                                                                                                  \\
			\hline \multicolumn{4}{|c|}{CNN}                                         & \multicolumn{4}{|c|}{FoxNews}                                     & \multicolumn{4}{|c|}{CNN}                       & \multicolumn{4}{|c|}{FoxNews}                                                                                                                                                                                                                                                                                                                                                \\
			\hline \multicolumn{2}{|c|}{TV}                                & \multicolumn{2}{|c|}{Twitter (+3 months) }                        & \multicolumn{2}{|c|}{TV}              & \multicolumn{2}{|c|}{Twitter (+3 months) } & \multicolumn{2}{|c|}{TV (+3 months) } & \multicolumn{2}{|c|}{Twitter}              & \multicolumn{2}{|c|}{TV (+3 months) } & \multicolumn{2}{|c|}{Twitter}                                                                                                                                                  \\
			\hline Token                                                             & Attr.                                                             & Token                                           & Attr.                                      & Token                                           & Attr.                                      & Token                                           & Attr.                                      & Token                  & Attr. & Token                  & Attr. & Token                 & Attr.  & Token                 & Attr.  \\
			\hline know                                                              & 0.33                                                             & \textbf{protests}                               & 0.24                                      & \textbf{right}                                  & -0.24                                     & Baptist                                         & -0.24                                     & \textbf{Americans}     & 0.25 & \textbf{Americans}     & 0.21 & \textbf{cities}       & -0.22 & \textbf{blacks}       & -0.18 \\
			\hline going                                                             & 0.32                                                             & \textbf{people}                                 & 0.24                                      & country                                         & -0.24                                     & moral                                           & -0.22                                     & \textbf{ movement}     & 0.20 & church                 & 0.21 & \textbf{democrats}    & -0.20 & \textbf{cities}       & -0.18 \\
			\hline \textbf{people}                                                   & 0.31                                                             & technology                                      & 0.23                                      & lot                                             & -0.21                                     & \textbf{right}                                  & -0.21                                     & plaza                  & 0.21 & citizens               & 0.18 & \textbf{lives}        & -0.20 & \textbf{democrats}    & -0.20 \\
			\hline plaza                                                             & 0.28                                                             & county                                          & 0.21                                      & \textbf{violence}                               & -0.20                                     & toxic                                           & -0.18                                     & \textbf{president}     & 0.23 & \textbf{movement}      & 0.23 & mayor                 & -0.20 & federal               & -0.21 \\
			\hline \textbf{lives}                                                    & 0.27                                                             & \textbf{movement}                               & 0.21                                      & way                                             & -0.20                                     & praising                                        & -0.17                                     & \textbf{states}        & 0.21 & powerful               & 0.17 & \textbf{moral}        & -0.23 & Jackson               & -0.24 \\
			\hline that                                                              & 0.27                                                             & \textbf{lives}                                  & 0.20                                      & \textbf{white}                                  & -0.19                                     & \textbf{blacks}                                 & -0.17                                     & that                   & 0.21 & \textbf{president}     & 0.14 & officers              & -0.21 & liberals              & -0.19 \\
			\hline \textbf{movement}                                                 & 0.24                                                             & family                                          & 0.20                                      & \textbf{support}                                & -0.18                                     & \textbf{violence}                               & -0.16                                     & they                   & 0.21 & protests               & 0.20 & \textbf{organization} & -0.19 & \textbf{lives}        & -0.24 \\
			\hline \textbf{president}                                                & 0.22                                                             & \textbf{protesters}                             & 0.20                                      & \textbf{blacks}                                 & -0.18                                     & \textbf{support}                                & -0.15                                     & today                  & 0.20 & \textbf{states}        & 0.24 & party                 & -0.24 & \textbf{moral}        & -0.20 \\
			\hline \textbf{protesters}                                               & 0.18                                                             & virus                                           & 0.20                                      & want                                            & -0.18                                     & \textbf{white}                                  & -0.15                                     & \textbf{world}         & 0.21 & \textbf{world}         & 0.13 & \textbf{shoot}        & -0.18 & \textbf{organization} & -0.18 \\
			\hline \textbf{protests}                                                 & 0.18                                                             & \textbf{president}                              & 0.19                                      & homicides                                       & -0.17                                     & matters                                         & -0.13                                     & \textbf{year}          & 0.24 & \textbf{year}          & 0.17 & \textbf{cities}       & -0.22 & \textbf{shoot}        & -0.21 \\
			\hline
		\end{tabular}}

	\caption{Top 10 tokens most predictive of how CNN and Fox News TV stations and Twitter users replying to @CNN and @FoxNews use keywords topically related to \textit{Black Lives Matter}.}
	\label{table b-4}
\end{table*}
\clearpage
\clearpage
{\fontsize{9.1pt}{10.0pt} \selectfont \bibliography{aaai22}}
\end{document}